\documentclass[11pt,a4paper]{article}
\usepackage[hyperref]{acl2020}
\usepackage{times}
\usepackage{latexsym}
\usepackage{float}

\definecolor{mygray}{gray}{.9}

\usepackage{graphicx}
\usepackage{algorithm}
\usepackage{algorithmic}
\usepackage{hyperref}
\usepackage{booktabs}
\usepackage{url}
\usepackage{subcaption}
\usepackage{float}
\usepackage[export]{adjustbox}
\usepackage{colortbl}
\usepackage{multirow}
\usepackage{comment}
\usepackage{xcolor}

\usepackage{bbding}
\usepackage{pifont}
\usepackage{amssymb}
\usepackage{amsmath}

\usepackage{microtype}

\aclfinalcopy % Uncomment this line for the final submission
\newcommand{\modelname}{PRAL}
\newcommand{\tablestyle}[2]{\setlength{\tabcolsep}{#1}\renewcommand{\arraystretch}{#2}\centering\footnotesize}

\makeatletter\renewcommand\paragraph{\@startsection{paragraph}{4}{\z@}
  {.5em \@plus1ex \@minus.2ex}{-.5em}{\normalfont\normalsize\bfseries}}\makeatother

\title{A Tailored Pre-Training Model for Task-Oriented Dialog Generation}

\author{Jing Gu \quad Qingyang Wu \quad Chongruo Wu \quad Weiyan Shi \quad Zhou Yu \\
University of California, Davis \\
\texttt{\{jkgu, wilwu, crwu, wyshi, joyu\}@ucdavis.edu} \\
}
\date{}

\begin{document}
\maketitle

\begin{abstract}
% Recent reserachers have been focusing on applying pre-training in the dialog system. However, these models are not fit for the dialog systems. We propose Pretrained Role Alternating Language model (PRAL) that is designed for dialog system in its nature. 

The recent success of large pre-trained language models such as BERT and GPT-2 has suggested the effectiveness of incorporating language priors in down-stream dialog generation tasks. However, the performance of pre-trained models on dialog task is not as optimal as expected. 
In this paper, we propose a Pre-trained Role Alternating Language model (PRAL), designed specifically for task-oriented conversational systems. We adopt ARDM \cite{ARDM} that models two speakers separately. We also design several techniques, such as start position randomization, knowledge distillation and history discount to improve pre-training performance. We introduce a task-oriented dialog pretraining dataset by cleaning 13 existing data sets. We test \modelname{} on three different downstream tasks. The results show that \modelname{} performs better or on par with the state-of-the-art methods. 
% We theorize that there are two issues: these pre-trained dialog systems did minor changes on GPT-2, which were not designed for dialog datasets. Moreover, there are not enough corpora geared towards dialog datasets. 

\end{abstract}

\section{Introduction and Related Work}

The current approaches to build task-oriented dialog systems still require a substantial amount of annotations and therefore are labor-intensive. On the other hand, large-scale pre-trained language models such as BERT \cite{bert} and GPT \cite{gpt2} have achieved great success on various NLP tasks, which proves the effectiveness of pre-training. There have been several attempts to directly apply these language models to dialog systems. For example, Transfer-Transfo \cite{transfertransfo} fine-tuned GPT on the Persona-Chat dataset \citep{persona-chat-dataset} and achieved the state-of-the-art performance on chitchat dialog generation. %, which shows the importance of language model fine-tuning in dialog systems.  
\citet{first-task-oriented} adopted the structure of Transfe-Transfo, further pre-trained GPT-2 with a collection of task-oriented dialogs and obtained good results on downstream tasks. DialoGPT \cite{dialoGPT} utilizes a large Reddit corpus to further pre-train GPT-2 \cite{dialoGPT}. All of these studies pointed to a promising direction towards building dialog systems with large-scale language models and less supervision. %We also notice there is a concurrent work  \par
% Therefore, their results are usually discouraging compared with their performance on other generation tasks \citep{ARDM}

%Since these pretrained-models are not specifically targeted at dialogue system development,  pure language models are not able to reach good performance on dialog tasks due to the following reasons: 1) dialogues consists of multiple-parties and each party has different language styles --> two role \cite{ardm}. 2) catastrophic forgetting, knowledge base search --> knowledge distilation 3) stronger contextual information --> discounted factor. Therefore, we proposed the , which pretrained with high-quality dialogue dataset along with three  techniques to deal with these problems.

\begin{figure}[t]
%\small
\centering
\includegraphics[scale=0.42]{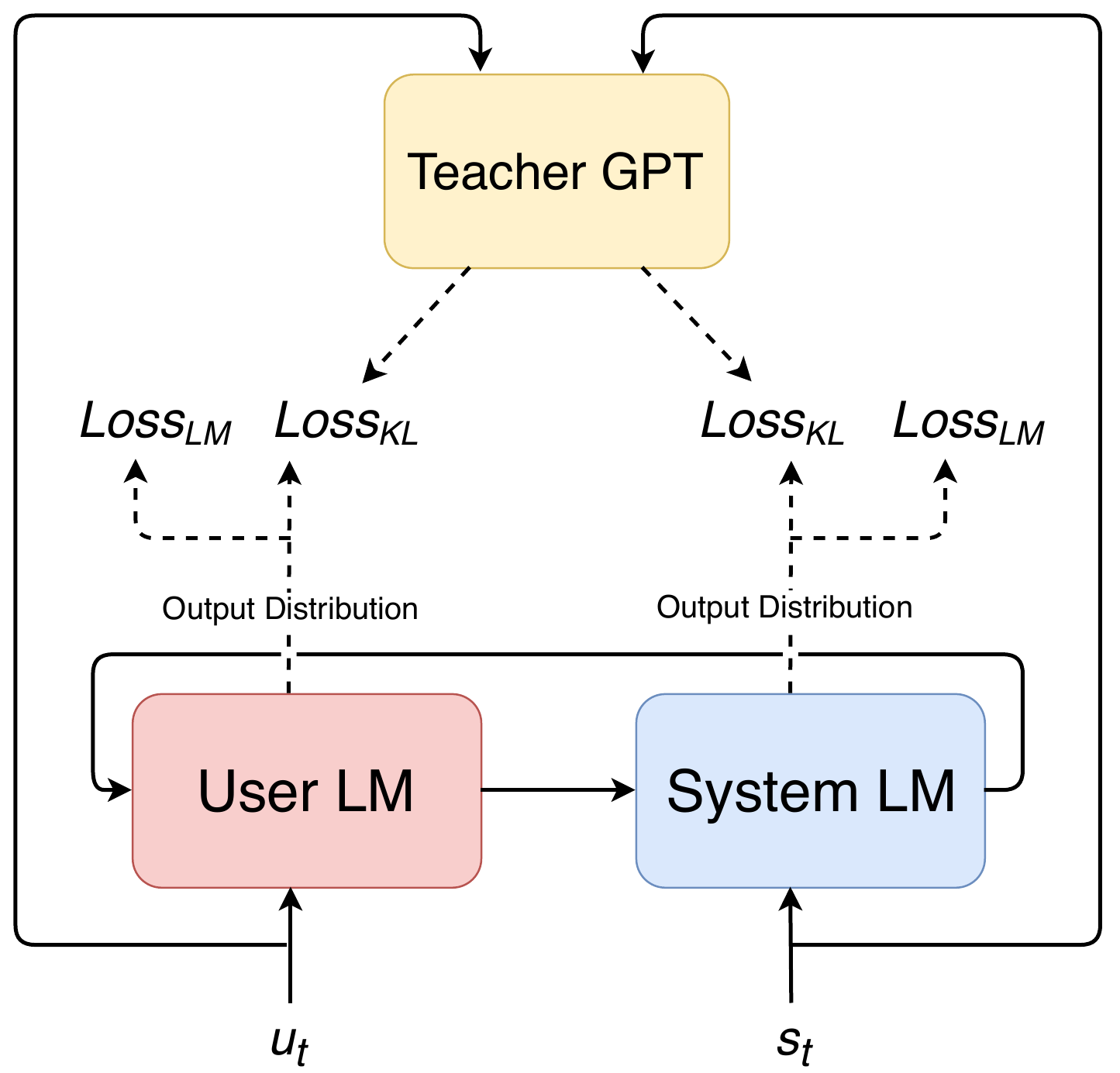}
\caption{The illustration of PRAL. There are two language models for the user and the system, respectively. Teacher GPT is used to provide a better supervision to them. $Loss_{LM} $ and $Loss_{KL}$ denote the losses for the language modeling and the KL divergence.}
\label{fig:model}
\end{figure}

However, these languages models applied on dialog systems still have some limitations.  %there are still two limitations on these existing dialog systems. 
First, further pretraining language models for dialog systems requires a huge amount of  training corpora, but a diverse collection of high-quality dialog datasets is always hard to obtain. Second, dialogs consist of multiple-parties and each party has different language styles. However, most previous dialog systems only utilize one single language model to perform the dialog generation for all parties. % and are not able to capture the language dynamics.
Next, dialogs are always of variable lengths, and therefore the fixed-length position embedding in GPT results in sub-optimal results. Additionally, dialogs involve a large amount of commonsense knowledge which can be missing in small-size language models. Furthermore, natural dialogs require good understanding of the context, yet contextual information is hard to preserve in language models.

%1) Most of the recent large pre-trained dialog system did minor change based on GPT. Since GPT is not designed specifically for dialog generation tasks, these models usually also can not fit perfectly for dialog generation tasks. 
% First, most of the recent large pre-trained dialog system is based on GPT. However
%2) Since the high qualities dialog dataset is relatively scarce than general text, the current large pre-trained language model suffers from training resources shortage. 
% it is also hard to find a suitable dataset to pre-train a large language model with good performance.
% However, one major limitation of these pre-trained large dialog system is that they use only one language model, which is hard to model to model the distribution of responses by different roles in a conversation. ARDM \citep{ARDM} proposed to use different language models to simulate different roles respectively in dialog system and achieved state-of-the-art performance on differenet dialog tasks without annotation.

 To tackle these issues, we propose Pre-trained Role Alternating Language model (\modelname{}), a language model specifically designed for dialog generation. To begin with, we collect and process 13 dialog datasets, ranging from TV transcripts to pizza ordering dialogs, to enrich the pretraining data with high-quality dialog corpora. Second, we adopt ARDM proposed in \citet{ARDM} and use two separate GPT-2 to model the two speakers in the dialog.  Next, we apply Start Position Randomization (SPR) to cope with the variable lengths in dialogs, which also prevents the language model from binding the position index with the text information. Additionally, we utilize the original large-scale GPT-2 to perform knowledge distillation and incorporate common sense knowledge into the dialog generation. Finally, we re-weight each utterance with discount factors and emphasize on the later part in a dialog to better incorporate contextual information.% since the later part usually contain more condensed %aggregate more
 %contextual information.
 %Given the fact that the size of tokens in most dialogs is short than the maximum position index 1024, we also designed Random Start Position (RSP) to make each position embedding get a fair chance to update. RSP also prevents \modelname{} from binding position index with text. 
% First, since the original GPT-2 contains abundant knowledge, we use another frozen large GPT-2 as a constraint to reduce the probability of catastrophic forgetting. Meanwhile the frozen GPT-2 will also serve as the source of knowledge distillation. 
 % since latter utterances in a dialog contain more contextual information and are valuable for training, we use discount factor to make \modelname{} to pay more attention on the latter context. 
 We evaluate \modelname{} on three task-oriented datasets (CamRest676, Multiwoz and PersuasionForGood), and reach the state-of-the-art results without using any annotation. %It is worth noting that \modelname{} does not require annotation in evaluation process. In CamRest676 dataset, it achieve state-of-the-art results. While on Multiwoz and PersuasionForGood tasks, \modelname{} outperforms or is on par with state-of-the-art methods. 

In summary, we process and present a collection of high-quality dialog datasets suitable for pre-training large-scale language models on dialog systems.
We also propose \modelname{} and design several effective techniques to improve the dialog model pretraining. Our pretrained model leads to an increase on success rate on CamRest676 and MultiWOZ dataset, and an improvement on the  coherence and diversity scores by 50\% on  PersuasionForGood.

\begin{table}[t]
%\centering
%\small
%$\setlength{\tabcolsep}{100pt}
\tablestyle{15pt}{1.08}`
\begin{tabular}{lc}
 \toprule
 \textbf{Dataset Statistics}                             &  %\textbf{Our Dataset}
 \\ \hline
  \# Domains & 13\\
\# Dialogues                                 &  142,298 \\
 %Total \# turns                              &  901,363  \\
 %Total \# words                              &  25,088,854 \\
% Avg. length(\#char) per dialogue            &  745.37 \\
 Avg. turns per dialogue                     &  12.66 \\
 Avg. tokens per turn                        &  11.78  \\
 Avg. tokens per dialogue                    &  149.25 \\
 Total unique tokens                         &  108,106\\
 \bottomrule
\end{tabular}
\caption{Statistics of our dataset}

\label{table:dialog-statistic}
\end{table}

\section{\textit{PretrainDial} Dataset for Pretraining}

%To our best knowledge, there exists no existing large dataset corpus
Clean dialog datasets that are big enough to pre-train language models for dialog systems are difficult to find. Therefore, we propose \textit{PretrainDial}, a large-scale multi-domain dialog corpus suitable for pretraining. We carefully selected 13 existing dialog corpora listed in Appendix~\ref{append: dataset_soures}, ranging from chitchat such TV transcripts to task-oriented dialogs, and process them in a unified form. Table. ~\ref{table:dialog-statistic} shows the statistics of \textit{PretrainDial}.

\section{Methods}
\label{sec:methods}

We adopt the architecture from ``Alternating Roles Dialog Model'' (ARDM) ~\cite{ARDM} which uses two language models for the user and system separately. Each language model is initialized with a small GPT-2 \cite{gpt2}. In this section, we will briefly introduce ARDM and describe our approaches to improving existing language models. Figure \ref{fig:model} shows the main structure of \modelname{}  
\subsection{Alternating Roles Dialog Model}
We first briefly talk about Alternating Roles Dialog Model (ARDM)\cite{ARDM}. The basic idea behind ARDM is to simultaneously model the user and system with two separate GPT-2 to capture the different language styles. A dialog can be considered as a sequence of utterances $d=\{u_1, s_1, u_2, s_2, \dots, u_T, s_T\}$, where $T$ is the total number of turns. We use $p_{u}$ and $p_{s}$ to represent the probability of the user utterance and system utterance. The entire dialog distribution is defined as: 

\begin{equation}
    p(d) = \prod_{t=1}^T p_u(u_t | u_{<t}, s_{<t}) \, p_s(s_t | u_{\leq t}, s_{<t})
    \label{eq:lm}
\end{equation}

% $p_u$ and $p_{s}$ can be calculated with the user and system  language models respectively. For an utterance $u_t$ or $s_t$ with $m$ tokens $\{w_1, \dots, w_m\}$, the joint probability of an utterance will be

% \begin{equation}
%     p_u(u_t | u_{<t}, s_{<t}) = \prod^{m_{u_t}}_{i=1} P(w_i | w_{< i}, u_{<t}, s_{<t})
%     \label{eq:lm_user}
% \end{equation}
% \begin{equation}
%     p_s(s_t | u_{\leq t}, s_{<t}) = \prod^{m_{s_t}}_{i=1} P(w_i | w_{< i}, u_{\leq t}, s_{<t})
%     \label{eq:lm_sys}
% \end{equation}
% Finally, ARDM train the dialog model by maximizing the likelihood over Equation 1.
By maximizing the likelihood in Equation~(\ref{eq:lm}), ARDM successfully models the user and system at the same time. However, ARDM did not employ additional pre-training on the dialog corpus. 
In contrast, we further pre-train ARDM on our collected dialog corpus.
In addition, we propose three effective techniques to help pre-training.

\subsection{Start Position Randomization}

% Since all the tokens of one sentence are fed into GPT2 in parallel, GPT2 use position index to label positional relation of different input tokens. However, in most cases, the length of input tokens is shorter than the maximum index 1024. As a result, the latter positions get less trained. As a result, the performance of GPT2 will performance not so ideal if the generated sentence is long. Besides, if the position start index is fixed, the model will also learn to generate responses based on position rather than current word embedding or history.

We use GPT-2 as the language model in \modelname{}. GPT-2 uses position embedding to encode the location information for each token. It supports the maximum position of 1024, and the position index always starts from 0. However, since most dialogs contain less than 1024 tokens, most vectors in the positional embedding would remain zero and not be updated during pre-training. Besides, since position embedding only provides the location information for each token, fixing the start position to 0 will bond certain text with certain position index. For example, ``hi'' is always bonded with index 1 as ``hi'' usually appears at the beginning. The model is likely to overfit on the positional embeddings near the start. 

To address these issues, we propose to use Start Position Randomization (SPR). Denoting $L$ as the total number of tokens in a dialog, then the maximum start position index is $1024-L$. We randomize the start position to be any number between 0 to $1024-L$. It would disentangle the positional information from the textual meaning and force the model to update all the positional embeddings.

%As the original positional embedding in GPT-2 contains 1024 different vectors,
%However, since most utterances in a dialog are much shorter than 1024 tokens, if the dialog always starts at position zero, most of the positional embedding will not be updated. 
% Besides, since the positional embeddings should only provide the location information of each token, fixing the start position to 0 will bond certain text with certain position index, which misleads the model to fix the position embedding text meaning.
% As a result, the model is likely to overfit on position embedding. By SPR, all positional embeddings are given an equal chance to be trained. 

\subsection{Teacher GPT}

% A commonly used technique in training the language model is label smoothing. However, it usually hurts preserving the language model distribution pre-trained on large corpus \cite{label-smoothing}.
% To alleviate this problem, instead of using traditional label smoothing where the target token is one-hot vector, 

All neural networks suffer from the catastrophic forgetting problem \citep{overcome_catastrophic_forgetting}. Since we have trained GPT-2 with the new dialog corpus and obtained a new language model, the new model is at risk in forgetting the prior knowledge from the original GPT-2.%it's still possible that the new model % pretrain  the original GPT-2  on dialog corpus and up initialize \modelname{} from the pre-trained version by \cite{gpt2}, when  train \modelname{} on dialog corpus, it still risks of forgetting the knowledge from its original pre-training corpus.
 %\modelname{} utilizes the weights in the original GPT-2. 
Therefore, we apply a simple approach as continual learning \cite{copntinual-learning} to mitigate the problem. 
% In more details, we freeze another GPT-2 as the model for continual learning.
In detail, we use another fixed GPT-2 as the teacher network to preserve the knowledge.
% Training data will be input to both the language models and the frozen GPT-2 at the same time. 
To do so, we use the distillation loss \cite{knowledge-distillation} which calculates the KL divergence between our model and the fixed GPT-2, $\text{KL}(p,\;p^{constriant})$:

\begin{comment}
\begin{equation}
Loss_{KL} = \text{KL}(\; p^{gpt}, \,p\;)
\end{equation}
\end{comment}

In our best model, we use GPT-2 large as the teacher language model to distill more knowledge. 
% Since the large model incorporates more knowledge than the small one, it can be considered as a form of knowledge distillation . 
Because applying a larger GPT-2 requires more computational resources, we also conduct the ablation of using GPT-2 small as the teacher language model in the experiments. The result suggests that regardless of the size of the GPT-2, our method helps in the dialog model pretraining process.

\subsection{History Discount}
\label{subsection: discount_factor}

In each dialog, utterances in the latter part should have more importance because they aggregate more complex contextual information, which can help the model to learn the consistency in context. Therefore, we introduce discount factor $\gamma$ to re-weight the importance of each utterance based on the turn number. For a dialog with a total of $U$ utterances and the current utterance index $u$, the language model loss is weighted by $\gamma^{U-u}$. By multiplying the discount factor $\gamma$, the model has stronger ability to predict complex context and generate more consistent responses.

\subsection{Optimization}
%To have generated sentences in high quality, 
We use the loss for language modeling to optimize the model, as shown below in Equation \ref{eq:lm}, %\ref{eq:lm_user} and \ref{eq:lm_sys}% 
%We could get $Loss_{LM}$:

\begin{equation}
Loss_{LM} = \sum_{u=1}^{U}{ \,
                \gamma^{U-u}
                \sum_{l=1}^{L_u -1}
                CE(P_{ul}, \,G_{u(l+1)}) 
            }
\end{equation}

 CE here denotes cross entropy loss. $U$ is the total number of utterance in a dialogue, and $L_u$ is the total number of tokens in the $u^{th}$ utterance. For the loss of each utterance $u$ in the dialogue, it is weighted by the discount factor described in section~\ref{subsection: discount_factor}. We go over each word in the utterance, except for the last one, to compute its cross-entropy loss between the output probability distribution $P_{t(l+1)}$ and its ground truth $G_{t(l+1)}$.

% During training period, when we calculate the language model loss for a utterance, we also compute the KL loss for each word in the dialogue 

% \begin{equation}
% \small
% Loss_{KL} = \sum_{u=1}^{U}{ \,
%                 \sum_{l=1}^{L_u}
%                 KL(P_{ul}, \,P_{ul}^{gpt}) 
%             }
% \end{equation}

Our final loss will be a combination of the language model loss and KL divergence: 
\begin{equation}
Loss = Loss_{LM} + \alpha \, \text{KL}(p,\;p^{constriant})
\end{equation}

The factor $\alpha$ is used for better optimization and will be decreasing exponentially as the number of iteration increases, i.e. $ \alpha = \alpha_0 \, \lambda^{iter}$.

\begin{table*}[h]
%\centering

%\hspace{-1mm}
\begin{subfigure}[t]{0.39\textwidth}
%\centering
\tablestyle{3.1pt}{1.12}
\begin{tabular}{lcc}
 \toprule
 \textbf{Model} & \textbf{BLEU-4} & \textbf{Success F1}  \\ \hline
 Sequicity & 21.4 & 0.852 \\
 Sequicity (w/o RL) & 22.9 & 0.821 \\
 GPT-2-finetune  & 21.8 & 0.851 \\
 DialoGPT & 25.2  & 0.861 \\
 ARDM  & 26.2 & 0.864  \\ \hline
 \modelname{} & \textbf{27.3} & \textbf{0.870}\\
 \hspace{2mm} - w/ Teacher GPT(small) & 26.9 & 0.869  \\
 \hspace{2mm} - w/o Teacher GPT & 25.0 & 0.865  \\
 \hspace{2mm} - w/o loss discount & 27.0 & 0.867 \\
 \hspace{2mm} - w/o SPR & 26.6 & 0.869 \\
 \bottomrule
\end{tabular}
\caption{Results on CamRest676 dataset.}
\label{tab:camres}
\end{subfigure}
\,\,\,
\begin{subfigure}[t]{0.58\textwidth}
%\centering
\tablestyle{3.5pt}{1.53}
\begin{tabular}{l|cc|ccc}
 \toprule
 \textbf{\multirow{2}*{Model}} & \multicolumn{2}{c|}{\textbf{Supervision}} &  \textbf{\multirow{2}*{BLEU-4}} & \textbf{\multirow{2}*{Inform}}  & \textbf{\multirow{2}*{Success}}  \\ 
 & Dialog State & Dialog Act &&& \\
 \hline 
 Human & - & - & -  & 0.989 & 0.965 \\ \hline 
 Baseline & $\checkmark$ &  $\times$ & 18.9  & 0.825  & 0.729  \\
 HDSA &  $\checkmark$ &  $\checkmark$ & \textbf{23.6}  & \textbf{0.877} &  0.734 \\
 LaRL &  $\checkmark$ &  $\times$ & 12.8  & 0.828  & \textbf{0.792}  \\ \hline
 ARDM &  $\times$ &  $\times$ & 20.6 &  0.874 &  0.728 \\
 \modelname{} &  $\times$ &  $\times$ & 21.6  & 0.875  & 0.742  \\
 \bottomrule
\end{tabular}
\caption{Results on MultiWOZ dataset}
\label{table:mutliWOZ}
\end{subfigure}

\begin{subfigure}[t]{0.95\textwidth}
%\centering
\tablestyle{3.1pt}{1.05}
\begin{tabular}{l|ccc|cccccc}
 \toprule
      & Perplexity $\downarrow$  & BLEU-1 $\uparrow$   &  BLEU-2 $\uparrow$  &  Fluency $\uparrow$  & Logic $\uparrow$   & Coherence $\uparrow$   & Diversity $\uparrow$  & Overall $\uparrow$  & Avg.Donation $\uparrow$
      \\ \hline
 ARDM & \textbf{10.1}  & 16.5  & 6.44  & 0.39  & 0.41  & 0.37 & 0.27 & 0.18 & 0.62  \\
 \modelname{} & 10.3  & \textbf{17.3}
 & \textbf{10.9}  & \textbf{0.61}  & \textbf{0.59}  & \textbf{0.63} & \textbf{0.73} & \textbf{0.82} & \textbf{0.99}  \\
 \bottomrule
\end{tabular}
\caption{PersuasionforGood. Automatic Evaluation and Human Evaluation Results}
\label{table:persuasion}
\end{subfigure}

\caption{Evaluation on three datasets} 
\end{table*}

\section{Experiments}

We pre-train \modelname{} on \textit{PretrainDial}. For the pre-training detail, please refer to Appendix~\ref{append: Training Details}. %\par
%\subsection{Training Details}
% \subsection{Dialog Generation Task Evaluation}
To show the generalizability of \modelname{}, we evaluate it on three task-oriented dialog tasks, CamRest676, MultiWOZ and PersuasionforGood.

\noindent \textbf{CamRest676}~\cite{RojasBarahona2016ANE} is a small dialog dataset for restaurant recommendation in  Cambridge. There are 680 dialogues where users look for restaurants based on their preference on food, price range and area.
Table.~\ref{tab:camres} shows our results on CamRest676. We use BLEU-4 metrics to measure the quality of generated sentences, and Success F1 to evaluate the responses on specific slots, such as address, phone, postcode. Sequicity is the state-of-the-art method in task-oriented dialog tasks that utilizes annotations in a traditional fashion. We found that \modelname{} is able to beat all the baselines on both BLEU-4 and Success F1 including the state-of-the-art ARDM model. One thing to note is that \modelname{} doesn't need any annotation. This suggests that  \modelname{}  leverages external knowledge from the  pre-training process, and the proposed techniques are effective for dialog language model pretraining. %\modelname{} outperforms the state-of-the-art method ARDM in both BLEU-4 and Success F1. \par

We also perform ablation study on CamRest676 and find that the Teacher GPT plays the most important role. This suggests  knowledge distillation from the large pre-training is critical to good performance. Our model also outperforms the  DialoGPT baseline, which  utilizes a much larger Reddit dataset (30G) in pretraining compared to the much smaller but higher-quality \textit{PretrainDial} (300MB) data used in \modelname{} . This suggests the quality rather than the size of the dataset matters.  

\noindent\textbf{MultiWOZ} \cite{Multiwoz} is a large-scale multi-domain dataset, which contains around 10k dialogues covering various domains. We evaluate the models with on BLEU-4, Inform Rate and Success Rate which measures if the system provides the requested information. Table.~\ref{table:mutliWOZ} shows our results.
We first compare our model to the attention seq2seq model used as the baseline in Multiwoz \citep{Multiwoz}.
We then compare our model with HDSA \citep{Multiwoz} and LaRL \cite{larl-multiwoz-baseline}. Our model outperforms or achieve comparable results with HDSA and LaRL. \modelname{} achieves a much higher BLUE-4 score than LaRL (improve 68.8\%). \modelname{} outperforms ARDM in all metrics. It is worth noting our model does not use any annotation.
% This shows that pretraining on dialog corpura is effective on task-oriented dialog tasks. 
%\input{figures/res_table_MultiWOZ.tex}

\noindent\textbf{PersuasionforGood}
%\subsubsection{PersuasionforGood}
%\input{figures/res_table_persuasion.tex}.
We also evaluate our method on a non-collaborative dialog dataset, Persuasion for good~\cite{Wang2019PersuasionFG}. In PersuasionforGood, a persuader tries to persuade another user to donate money. There are a total of 1,017 dialogues.
Unlike CamRest676 and Multiwoz, the language in PersuasionforGood dataset is so diverse that  BLEU-4 scores of all of the models on PersuasionforGood are too low to be a scientific metrics. 
Therefore, we use BLEU-1 and BLEU-2 instead. Compared with ARDM, our model achieves a significant higher score on BLUE metrics, especially on BLEU-2 (63\% up). We also conduct human evaluation between ARDM and our model. We ask human evaluator that how much they are willing to donate after the conversation and acquire their ratings on the dialog system in terms of fluency, logic, coherence and diversity.
% Then they will be asked which one performs better in term of fluency, logic, coherence, diversity, and overall preference. 
The result of human evaluation suggests that \modelname{} outperforms ARDM on all the metrics and is a better language model for dialog system in general. 
%Table.~\ref{table:persuasion} shows the automatic metrics results and human evaluation results.
For examples of the persuasion process, please refer to Appendices \ref{append: dialog example}.

\section{Conclusion}
We propose \modelname{}, a large pre-trained language model for task-oriented dialog systems. We successfully incorporated methods that are designed for large pre-trained language models into \modelname{} and achieved good performances on three downstream tasks. Specifically, we designed  start position randomization, knowledge distillation and history discount to improve pre-training performance. The model generates more fluent, coherent, diverse and logical dialogs according to human evaluation results. The resulting dialog systems also obtained more donation.
We also clean a high quality dialog dataset for pre-training process. Our work is the first step towards a coherent and engaging dialog model that generalize to different dialog tasks.%The results on three dialog tasks indicate the effectiveness of our proposed method.
% However, we should be aware of ethical concerns of the misuse of our model.
% on a persuasion task

%\bibliography{anthology,acl2020}
\bibliography{acl2020}
\bibliographystyle{acl_natbib}

\clearpage
%\newpage
\appendix

\section{Appendices}
\label{sec:appendix}

\subsection{Training Details}
\label{append: Training Details}
We adopt the architecture from ARDM by using two language model to simulate the user and the system. For the language models, we adopt pre-trained language model GPT-2 small \cite{gpt2}. For teacher neural model, we use GPT-2 large \cite{gpt2}.
We follow the same special format in GPT-2 as the ``trigger" so the model can zero-shot dialog response. In detail, we use ``A:" and ``B:" as user role prefix and use ``\textbackslash{}n\textbackslash{}n\textbackslash{}n" as suffix. 
We use AdamW optimizer. The number of warm-up steps is set to be 10 percent of the total training step. The learning rate is set to be $1 \times 10^{-4}$. 
For the calculation of loss, we set $\alpha_{0}$ to be 0.1 and set $\lambda$ to be 0.9999. The discount factor $\gamma$ is set to be 0.95.

\subsection{Dataset sources}
\label{append: dataset_soures}
Our dataset contains high-quality dialogues which are selected from other 13 datasets listed in Table~\ref{tab:13_dataset_source}.

\begin{table*}[h]
\onecolumn
\centering
\begin{tabular}{cc}
\toprule
 1 &  CCPE-M dataset~\cite{Radlinski2019CoachedCP}  \\ 
 2 &  ChitChat-Dataset~\cite{chitchat}  \\ 
 3 &  Conversations Gone Awry Dataset(Wiki)~\cite{Zhang2018ConversationsGA}   \\ 
 4 &  Conversations Gone Awry Dataset(CMV)~\cite{Zhang2018ConversationsGA}  \\ 
 5 &  Cornell Movie-Dialogs Corpus~\cite{DanescuNiculescuMizil2011ChameleonsII} \\ 
 6 &  DailyDialog~\cite{Li2017DailyDialogAM}    \\ 
 7 &  Frames Dataset~\cite{Asri2017FramesAC} \\ 
 8 &  PersonaChat ConvAI2 Dataset ~\cite{persona-chat-dataset} \\ 
 9 &  The Schema-Guided Dialogue Dataset~\cite{Rastogi2019TowardsSM} \\ 
 10 & Self-dialogue Corpus~\cite{Fainberg2018TalkingTM}  \\ 
 11 & Friends Series Dataset~\cite{friends_corpus}  \\ 
 12 & Taskmaster-1~\cite{Byrne2019Taskmaster1TA}   \\ 
 13 & Reddit Corpus~\cite{reddit_corpus}   \\
\bottomrule
\end{tabular}
\caption{13 Dataset Sources}
\label{tab:13_dataset_source}
\end{table*}

\clearpage
\subsection{Persuasion Example}
\label{append: dialog example}
One dialogue example is shown in Table~\ref{tab:human_machine_example}.

\begin{table*}[h]
    \centering
    \resizebox{1.0\textwidth}{!}{
    \begin{tabular}{c|p{14cm}}
    \toprule
    \textbf{Role} & \hspace{5cm}\textbf{Utterance} \\
    \hline
    \rowcolor{mygray}
    SYS & Good morning, how are you doing today? \\
    USR & Good! \\
    \rowcolor{mygray}
    SYS & Have you ever heard of Save the Children? \\
    USR & I have not heard of it. Would you like to give an introduction? \\
    \rowcolor{mygray}
    SYS & It is a charity that helps children in need around the world. \\
    USR & Cool \\
    \rowcolor{mygray}
    SYS & I would like to give a little bit more information. \\
    USR & Sure. Go ahead.\\
    \rowcolor{mygray}
    \multirow{2}*{SYS} & Save the Children is an international non-governmental organization that promotes children's rights, provides relief and helps support children in developing countries.\\
    USR & It is a good organization.  I am glad to know it. \\ 
    \rowcolor{mygray}
    \multirow{2}*{SYS}&Yes, they are a very professional organization. Would you be interested in donating some of your task money to this charity? \\
    USR & Sure. I would like to donate some money to it. \\
    \rowcolor{mygray}
    SYS& Thank you so much!\\
    USR& no problem. \\
    \rowcolor{mygray}
    \multirow{2}*{SYS} & How much do you like to donate to the charity now? Your donation will be directly deducted from your task payment. \\
    USR&  I would like to give \$2 to them. \\ 
    \rowcolor{mygray}
    SYS&  Thank you so much! \\
    USR& You are welcome \\ 
    \rowcolor{mygray}
    SYS& I hope you have a great day! \\ 
    USR& You too.\\ 
    \bottomrule
    \end{tabular}
    }
    \caption{An example conversation between a machine persuader trained by \modelname{} and a human persuadee.}
    \label{tab:human_machine_example}
\end{table*}

%\section{Supplemental Material}
%\label{sec:supplemental}

\end{document}